Penultimate draft. Accepted for publication in *Studies in the History and Philosophy of Science*

# WHAT CAN ARTIFICIAL INTELLIGENCE LEARN FROM MEDICINE? GENERATIVE ANALOGIES AND RELIABLE MACHINE LEARNING SYSTEMS

Emanuele Ratti[1], Department of Philosophy, University of Bristol

Lena Zuchowski, Department of Philosophy, University of Bristol

**Abstract.** In the past few years, machine learning (ML) has been widely (and to an extent, successfully) implemented in medicine. However, uncertainties surrounding ML have made it difficult to establish the bases of its epistemic and methodological warrants. In the literature, a parallel has been drawn between medicine and ML, suggesting that we should model epistemic and methodological standards for ML on the standards of clinical translation. By developing tools from Hesse's work, we characterise the nature of this parallel as a generative analogy between the process of clinical translation and the process of building ML systems. We identify more precisely the epistemic and methodological warrants of clinical translation that are typically only mentioned when appealing to the analogy, and we show in which sense such warrants apply analogically to the context of ML. In particular, we interpret warrants of clinical translation in reliabilist terms, and we show how this can inform a new form of ML reliabilism, which is distinct from (though compatible with) existing reliabilist accounts in philosophy of AI.

## 1 INTRODUCTION

In the past few years, philosophers and machine learning[2] (ML) practitioners alike have discussed to what extent we can rely on (or trust[3]) complex computational systems like ML tools, in particular, given the (ineliminable or essential) opacity and the uncertainties characterizing them (Alvarado and  2017). This question has been posed especially in the context of medical ML. In an influential article, London (2019) has proposed to treat these predicaments in the same way similar problems are handled in medicine itself. Medical knowledge itself is severely fragmented, and it can be characterized as "atheoretic, associationist, and opaque" (London 2019, p. 17). A paradigmatic case is our knowledge of pharmaceuticals, which is characterized by multiple unknowns, such as the uncertainties regarding mechanisms of action[4] or lack of theoretical

---

[1] mnl.ratti@gmail.com

[2] We use the terms AI and ML interchangeably. This might look like a controversial choice. Historically, the term 'AI' includes different approaches such as symbolic approaches often called 'good old-fashioned AI' (GOFAI), as well as ML approaches. However, it should be noted that within the context where the discussions of this paper take place, only ML is mentioned (and that is because of its characteristics, as we will see below), and AI is often used to refer to ML techniques. Therefore, within the context of this discussion it is not that strange to use AI and ML interchangeably.

[3] For the present article, it does not matter whether one is talking about trust or reliability, as discussions on trust assume reliability of ML tools as a precondition.

[4] While, at first sight, mechanisms of action are not essentially opaque (Humphreys 2011), one can make the argument that a given mechanism of action can be incredibly complex, and that it is unlikely that a human agent,

justification. But in medicine these unknowns are sidestepped by methodologies that have been proved effective in establishing reliable results. Given that with medical ML (and ML in general) we are facing a similar situation of atheoreticity, associationism and opacity, he draws a parallel between the warrants given by methods used in clinical translations with our prospects of evaluating ML models/systems (London 2019). This parallel is not limited to London's work (Russo 2023), and its popularity can be appreciated in the explosion of calls for more randomized controlled trials (RCTs) for medical AI, which is an implicit recognition that we should apply to ML roughly the same epistemic standards we apply to clinical translation (Genin and Grote 2021). But although this parallel can potentially be articulated into a concrete guide to overcome significant challenges raised by opacity or lack of methodological standards in ML, it has not gone beyond the level of a fascinating suggestion.

In this article, we analyse this parallel by interpreting it as a *generative analogy*, showing how the framework for evaluating the process of clinical translation can inform ML evaluation. In Section 2, we clarify what we mean by 'generative analogy' by constructing an account based on Hesse's work (1966). In Section 3, we show that the analogy is generative (3.1 and 3.2), and we clarify the aspects of clinical translation that are, indeed, generative (3.3). These are the mechanisms/processes for establishing the epistemic and methodological warrants of clinical translation, which we interpret in reliabilist terms. The scaffolding of these processes is what will inspire (i.e., it is generative of) an analogous framework for ML. In Section 4, we systematically investigate to what extent such a framework for clinical translation is transferable also in the context of ML, and how the analogy can lead to the formulation of epistemic and methodological warrants for the ML context that are compatible with reliabilist accounts already discussed in the literature (Duran 2026).

---

*qua* cognitive agent, will ever fully understand it through just a mechanism sketch or schema. It is also not unreasonable to compare the complexity of the mechanism of action of a molecule (involving millions of entities and activities) with the complexity of a ML model or the process of algorithmic optimization. So, it is possible to say that, at least in practice, mechanisms of action are essentially epistemically opaque.

## 2. GENERATIVE ANALOGIES

In this section, we provide a brief introduction to the Hessian account of analogies in science (section 2.1) before further developing a hitherto somewhat neglected kind of Hessian analogy, namely generative analogies (section 2.2).

### 2.1 Hessian Analogies in Science

An analogy is a comparison between two entities (the analogues), which are viewed as suitably similar such that this comparison can be used to draw one or more conclusions about the properties of the analogues. Hesse (1966) distinguishes between two kinds of analogies in scientific reasoning: a *positive analogy* exists between two analogues in virtue of some properties that they have in common, in contrast, a *negative analogy* exists if an analogue has one or more properties that the other does not, or if the same property is instantiated differently in the two analogues. Both positive and negative analogies can be used in scientific reasoning to form hypotheses about additional shared or non-shared properties of the analogues. In particular, if the properties in a positive analogy imply the existence of an additional property in one analogue, then one can deduce that (barring any screen factors) the second analogue should have this property as well. A classic example of such analogous reasoning is Reid's (1785) argument (reconstructed in Bartha, 2022) for life on Mars, which states that based on a number of similarities between our planet and Mars, it is "not unreasonable to think, that those planets may, like our earth, be the habitation of various orders of living creatures" (Reid, 1785, p. 24).

Hesse (1966) also provides a more nuanced analysis of analogical reasoning in terms of what she calls *vertical* and *horizontal relations*. Horizontal identity, difference or similarity relations are established between sets properties of two different analogues, while vertical identity, difference or similarity relations are established between properties of the same analogue. For the purpose of this paper, it is sufficient to review the former. Reid's (1785) reasoning about life on Mars is clearly an example of establishing horizontal relations between a set of properties (*p1*, *p2*, *p3*) on Earth and the same set of properties on Mars. In particular, the reasoning can then be paraphrased as follows: If Earth has properties (*p1*, *p2*, *p3*) and these properties cause life to flourish this planet (*p4*), and Mars likewise possesses properties (*p1, p2, p3*), then one is justified in drawing the inference p4 is possessed by Mars too, despite other differences between the planets. Hesse (1966) assigns such deductive reasoning from horizontal relations between positive analogies a *predictive purpose*: in general terms, if Analogue 1 has properties (*p1, p2, p3*) that strictly imply property *p4*, then, if Analogue 2 possesses the same properties (*p1, p2, p3*), one is warranted to predict that it also has property *p4*.

Prediction is not the only purpose that Hesse (1966) ascribes to reasoning from horizontal relations between analogues. Such reasoning can also have a *persuasive purpose*, which Hesse (1966) illustrates by referring to the analogy between the father-child and the state-citizen relationship, which is often used to highlight "the consequences, of a moral or normative character, which follows from the relations of the four terms already known" (Hesse 1966, p. 63), and thereby persuade people that this model is worth adopting.

But recognizing that reasoning from horizontal relations between positive analogues can have different purposes, opens the door to identifying further such purposes beyond the two recognized by Hesse (1966). In the following, we will argue that there is an important third purpose, namely, a *generative purpose* to analogous reasoning in science.

**2.2 Generative Analogies**

Consider the following well-studied episode from the history of molecular biology. By 1953, there were competing hypotheses on the mechanism behind protein synthesis, none of which involved nucleotides in any systematic way (Stegmann 2016). Watson and Crick's model of DNA (1953) inspired the physicist George Gamow (Kay 2000), who proposed to conceptualise the problem of the relation between DNA and amino acids purely as a coding problem, namely as the problem of finding the correct translation between a code with an alphabet of four letters (DNA bases), and a code with an  alphabet of twenty letters (amino acids). The analogy here is between the kind of problem the scientists set out to solve, rather than between sets of properties, as in the case of the predictive analogy discussed above. The outcome of this kind of analogical reasoning is not a prediction about the properties of the second analogue; rather, it is *generation of a research strategy*, which has been successful in the case of the first analogue and, due to the positive analogies between the analogues, one can now reasonably expect to be successful in the case of the second analogue as well. For the problem of finding the correct relation between DNA and amino acids, this generative analogy to coding problems proved successful: in their analysis of this episode, Kay (2000, p. 129) writes "to molecular biology the tropes of (...) information theory, linguistics, and computer-based cryptanalysis" and thereby mobilised additional research resources from outside the field. In particular, scientists began testing different coding schemes and testing them on state-of-the-art computational resources. The analogy was therefore truly generative in the sense that it generated both a viable, novel research strategy as well as additional research based on this strategy.

The example illustrates that generative analogies are successful analogies if they generate fruitful research strategies. As with predictive analogies, not every possible analogy will work. In the history of science we might find such failed analogies in the attempt to generate a research strategy for electrodynamics from an analogy between vortices and atoms, the now largely abandoned atomic vortex theory championed by Lord Kelvin in the 1860s (Falconer, 2019). In this case, the analogy captured some analogues features between the two fields but treating atoms as if they were vortices did not generate a fruitful research strategy. Without providing a full analysis of this case the crucial disanalogy between vortices and atoms seems to be the subatomic structure that simply cannot fruitfully be investigated by methods derived from vortex theory.

It should be noted that documented failed generative analogies are relatively rare: the fact that the atomic vortex theory is still of interest to philosophers of science is due to the high profile of some of its proponents. However, given that the outcome of a successful generative analogy is a fruitful research strategy, it makes sense that unsuccessful generative analogies leave few traces in the published record: scientists usually move on from unsuccessful research strategies without documenting failed attempts in detail.

We will argue in this paper that the parallel between ML and medicine is best viewed as a *generative analogy*. This means that the analogy does not predict properties; rather, it generates a research strategy. In section 3, we will argue for the interpretation of this analogy as a generative one. In section 4, we will outline the research strategy we think this analogy generates.

## 3. A NOVEL GENERATIVE ANALOGY BETWEEN CLINICAL TRANSLATION AND MACHINE LEARNING

As stated in the Introduction, parallels between ML and medicine have been drawn, with the goal of providing a more solid ground for the epistemology of ML by exploiting specific methodological moves used in medicine. Because what is done in medicine is envisioned to be done in ML, we interpret these parallels as attempts to draw some form of analogical reasoning. In the following we will briefly characterize the terms of the analogy (3.1), before showing why it is a generative, rather than a standard predictive one (3.2) . Finally, in 3.3 we identify which aspects of 'medicine' should be generative for ML, and then turn to Section 4 to see to what extent the analogy works for ML.

**3.1 The terms of the Analogy**

If we apply the received view on analogies as expressed by the Hessian predictive account to the comparison between the process of clinical translation and ML systems, this is what we have. The first analogue would be 'what is done in medicine'. In order to be aligned as much as possible to the context where the analogy was formulated the first time (London 2019), here 'what is done in medicine' is restricted to 'clinical translation', in particular in the way defined by London and Kimmelman (2015), and that will be specified below in detail. The second analogue is the construction of ML systems. In classic predictive setting, there will be a set of properties (*p1, p2, p3*, …, *pn*) that is shared by both, and then those properties will be used predictively (section 2.1). 'Use predictively' means to predict the existence of a fourth epistemic property that has already been established to predicate on the properties (*p1, p2, p3*) for clinical translation (i.e., the first analogue) and should therefore also do so for ML (i.e., the second analogue).

For example, London (2019) describes the process of clinical translation as characterised by the following three epistemic properties: associationism (p1), atheoreticity (p2) and opacity (p3). In particular, knowledge of pharmaceuticals and their effects proceed by providing evidence that strengthens the association between the use of a drug and a beneficial effect (p1); the establishment of this association does not crucially rely on theoretical knowledge (p2); and there is a pervasive lack of knowledge of mechanism of actions of many molecules which implies the impossibility of (mechanistically) explaining why they have the desired pharmaceutical effect (p3). The existence of these three epistemic properties is relatively unequivocally accepted (London 2019). These properties are seen as impediments to the mission of clinical translation. However, despite the fact that these seem to undermine the usual process of scientific discovery and justification, it is accepted that there are mechanisms (e.g. RCTs) by which the products of clinical translations are deemed reliable and the risks raised by p1-3 are managed (despite not solving those issues). This is seen as the fourth property (p4) in this scenario.

To establish the analogy, London (2019) then argues that ML systems possess the same three properties (p1, p2, p3). In particular, with respect to associationism (p1), he argues that ML tools do not track causal relations (with few exceptions), but patterns and regularities. With respect to atheoreticity (p2), it is a defining feature of ML that those tools learn from statistical associations between data sets rather than rely on explicitly coded theories (Termine et al 2026). With respect to opacity (p3), this is also a generally recognized and much discussed (e.g., Alvarado and Humphreys, 2017) feature of ML, whereby opacity applies both to the model that is generated as a result of training a ML algorithm, as well as to the process of optimization itself (Boge, 2022).

To add to London's considerations, it is important to show why having p1-3 raises challenges in ML. In fact, p1-3 reflects some troubling ways in which most ML systems operate (and their difficulties in adapting to evolving contexts), which can affect performance in unexpected ways (Freiesleben and Grote 2023; Grote et al 2024). For example, image classification tools based on deep neural network (DNN) in, e.g., the medical context (and others as well) are liable to the use of *shortcuts*, i.e., the establishment of associations based on non-medical information that has erroneously been encoded in the image and ultimately leads to misclassification of images not subject to the correlation exploited in the shortcut. This is because ML tools unveil associations between inputs and outputs, but these associations do not necessarily stand for more robust relations – there is always an uncertainty related to the associations that is difficult to quantify (p2). Due to the fact the DNN systems are opaque (p3) and do not rely on theories that would limit the scope of relevant associations (p1), such shortcuts are usually only detected once the algorithm starts misclassifying images. Another aspect affecting ML performance are *natural distribution shifts*, which is when a mismatch between deployment and training distribution causes a drop in performance. Grote and Freiesleben (2023) discuss an example based on COVID-19: imagine that a ML system has learnt to classify a person as having COVID-19 on the basis of certain symptoms (e.g., cough, fever, loss of sense of smell, etc). The problem is that the virus mutated rapidly, and such symptoms might not be as prevalent in COVID-19 cases after a few years. But we can anticipate the natural distribution shifts only if we know that the ML system indeed uses such symptoms to classify (which we might not, given p3), or if we have a way to connect the components of the model learnt by the algorithm to what we (theoretically) know about the disease (which we might not able to do, given p1). Similarly, in comparison to traditional image classification methods, ML image classification algorithms are more easily affected by *adversarial attacks*, i.e., deliberate tampering with the algorithm. In particular, due to the opacity (p3) of the algorithm, such attacks cannot be detected by tracking the internal workings of the process. Those examples seem to indicate that p1-3 are strictly connected to the specific ways in which ML often operates (shortcuts; natural distribution shifts; adversarial attacks), and they create, indeed, serious epistemic challenges.

**3.2 What Kind of Analogy?**

It can be argued that the analogy is predictive. In order to establish this, one needs to assume (i) that the fact that, e.g., RCTs are a suitable method of establishing the efficacy/effectiveness of drug treatments (p4) is predicated on the three relevant epistemic properties p1-3 and (ii) that both analogues possess those three properties. (ii) has been already established, while (i) deserves more attention. In fact, there are two problems with treating the analogy as predictive. Firstly, the analogy is not between two objects (e.g., Earth and Mars) but between two complex, epistemic processes, i.e., clinical translation and ML[5]. Similarly, the predicted property p4 is not a first order property of either analogue, but a process for testing the reliability of the two analogue processes. Second (and related to (ii)), even for the first analogue, it is not clear that the properties (p1, p2, p3) deductively entail the property p4. Instead, it seems to be more correct to say that empirical practice has proven that processes like RCTs can manage the risks raised by p1-3. Given that ML systems and their construction as a process have the same properties (p1-3), one can assume (but not deduce) that there is a high likelihood that a similar process to RCTs might also be effective in establishing the effectiveness of ML. However, this implies that the analogy is actually not a predictive one. Instead, the reformulation highlights the similarity between this analogy and the analogy between DNA sequences and coding problems (section 2.2).

We maintain that the best function to ascribe to the analogy between clinical translation and the construction of ML is a *generative* one. This means that the analogy generates a fruitful research strategy rather than specific predictions. In particular, it generates a research strategy that requires us to address the analogous challenges. This seems to remain in line with London's (2019) interpretation of the relationship between those two analogues: he suggests that we can learn a lot in the context of ML from the process of clinical translation, given that the process of clinical translation faces similar epistemic challenges as the construction and evaluation of ML systems.

---

[5] Admittedly, London (2016) himself remains somewhat ambiguous about this. On the one hand, he (2019) introduces the epistemic properties (p1, p2, p3) of medical knowledge by discussing drugs and pharmaceuticals as molecules; on the other hand, in other passages, he suggests that the analogy should be between the methods used to establish the reliability of the process of clinical translation and the methods used to validate or construct ML systems. Our unpacking of the analogy indicates that the latter is the intended meaning.

### 3.3 What Is It 'Generative' About Clinical Translation?

Now that we have established that the analogy is a generative one, we have to show which aspects of the first analogue are indeed 'generative'. By drawing from London's epistemology of clinical translation, we claim that the 'generative' aspects are the mechanisms/processes for establishing the epistemic and methodological warrants of pharmaceuticals, which we interpret in reliabilist terms. The scaffolding of these processes is what can inspire (i.e., it is generative of) an analogous framework for ML.

#### *3.3.1 The process of clinical translation*

Let us start by saying more about the process of clinical translation, and in which sense it can manage the risks and uncertainties raised in the context of medicine.

Clinical translation is defined by London and Kimmelman (2015; Kimmelman 2012) as a process through which information about the therapeutic potential of a molecule is accumulated - "the process of producing the information necessary to use a substance for therapeutic effect" (p 29). The process of clinical translation is typically associated with clinical trials. In fact, the structure of clinical translation is visualized as a complex system of trials (as shown in Figure 2 of their paper, p. 32). But trials are not all there is to clinical translation. In order to understand clinical translation, the starting point is that its output is not mere 'hardware' such as pharmaceuticals, devices, etc, nor is it just establishing a causal connection between an intervention and an endpoint. What is at stake in clinical translation is *information* - namely, information about the therapeutic potential of a molecule within given conditions and constraints; and where the potential is realised it is said that it has therapeutic efficacy or effectiveness[6] (we will refer to these as 'therapeutic usefulness', even if it might miss the nuances distinguishing the two terms). This is based on the assumption that "drugs alone are not therapeutic agents" (2015, p. 29). In fact, drugs are chemical or biological substances which can have a therapeutic effectiveness only in the presence of specific conditions and constraints. The process of clinical translation starts with a level of uncertainty where there is a hypothesis that a molecule might be therapeutically useful in treating a certain condition. But the hypothesis is, typically, based on either tenuous evidence or vague background knowledge: it may be that a molecule has a structure such that it is conceivable that its mechanism of action might be co-opted for therapeutic purposes; or it has been observed in other contexts that the molecule had provided benefits to treat a given condition; etc. In all these cases, the starting point of clinical translation is characterized by p1-3: there is uncertainty as to what the molecule does when administered in different physiological systems (opacity and atheoreticity); the association between the molecule and beneficial effects is tenuous or merely hypothetical (associationism). The process of clinical translation can then be seen as a mechanism to *manage* these uncertainties[7]. Most important, atheoreticity, associationism, and opacity are never fully overcome, but the risks they pose are better addressed with the right kind of information.

[6] Typically, therapeutic efficacy refers to what can be observed in ideal conditions of, say, a randomized controlled trials, while therapeutic effectiveness refers to effects in real-world conditions (see in particular Porzsolt et al 2015).

But what kind of mechanism is clinical translation, more precisely? According to Kimmelman and London, it is a mechanism for collecting and organizing pieces of information on the *molecule-in-context* that will show how the therapeutic potential of the molecule itself can be turned into therapeutic usefulness (if it can be therapeutically useful). The organized structured information about the molecule-in-context is called *intervention ensemble*. As ensembles, they have a complex structure. They are divided into three main dimensions, each of which has many components (Kimmelman 2012). Within the *treatment dimension*, components are typically information on dosage, schedule administration target, co-interventions, timing, risk mitigation, and others. The *population dimension* includes components such as information on diagnostic criteria, indications, contraindications, age. Finally, the *outcome dimension* includes components such as information on endpoints, duration, and the like.

In clinical translation, the values and boundaries of the components of these dimensions of intervention ensembles are explored, changed, and tested, with two possible outcomes. One outcome is that no adequate intervention ensemble is assembled and found, and this leads to the conclusion that the molecule does not have therapeutic usefulness (at least within the ensemble explored). A second possible outcome is that, in fact, the therapeutic potential can be realized. The potential is characterized in terms of two goals. First, the process of clinical translation has identified optimal values of variables within a given intervention ensemble (e.g. dose, timing of drug administration, etc) at which "a drug achieves the most favorable risk-benefit balance" (p 29). Second, the process has identified more precisely the boundaries of the dimensions outside of which the molecule is not therapeutically useful anymore.

[7] For an overview of quantitative and qualitative uncertainties in clinical research, see (Djulbegovic 2007)

To sum up, clinical translation can be construed as a process that, in an exploratory and piecemeal fashion, provides information on the conditions under which a certain molecule will unlock its therapeutic potential. These different pieces of information constitute an intervention ensemble. The more one departs from these ideal conditions, the less therapeutically useful the molecule. In the context of prescribing a drug, it is then a judgement-call by physicians to evaluate whether the context of a patient meets the characteristics of the relevant intervention ensemble so that a certain drug will be beneficial to the specific patient.

*3.3.2 Clinical translation as a reliable process*

Our takeaway from London and Kimmelman's discussion is that the process of clinical translation - that is, the process of assembling the right intervention ensemble – confers reliability to its outcomes in two ways. Before describing these senses, let us qualify more precisely the 'reliability' part.

We use reliability without appealing to the ways in which this term has been understood in classic forms in epistemology[8] (Alvarado and Humphreys 2017; Goldman 1979). Here we take the term 'reliable' in a broader sense: we use this term because we think that it captures best what we have in mind. The process of clinical translation - that is, the process of assembling intervention ensembles - is reliable if it produces a high ratio of accurate information about the ensembles that are explored. Here, accurate information means that it provides reasons to believe that an ensemble is therapeutically useful when, in fact, it is (i.e., when it unlocks the potential of a molecular when there is potential), or that it provides reasons to believe that the ensemble is not therapeutically useful when, in fact, it is not (i.e., it shows that the molecular does not have the potential, when it does not). This is the sense of 'reliability' of clinical translation procedures assembling intervention ensembles that, we think, correctly captures London and Kimmelman's views, when they say that "[e]fficient clinical translation therefore requires a process in which the effects of adjusting different dimensions can be sampled so that configurations that show signs of clinical promise can be identified and then promoted into further development" (p. 30). Some of these configurations will be further explored, others less so - in any case, clinical translation is reliable if it carries to the end of the process those intervention ensembles that are clinically effective, and blocks earlier intervention ensembles that are not. We can distinguish two types of reliability associated with this broad perspective.

[8] This claim must be qualified in two ways. First, despite our best efforts of agnosticism, our notion of reliability might partially overlap with Goldman-type reliabilism. Second, there is a parallel to be drawn with varieties of reliabilism in philosophy of computing, especially computational reliabilism (Duran and Formanek 2018). We will explore differences with computational reliabilism in the Conclusion.

In a first sense, a well-assembled intervention ensembles is reliable in producing a pharmaceutical doing something specific (e.g., treating a condition under specific/ideal circumstances). Second, an intervention ensemble provides information to reliably use the products of the process of clinical translation in new clinical contexts. The work here is to identify the conditions that make such a process reliable in these two senses.

We call the first sense of reliability *ensemble-reliability*. This means that the set of practices to assemble an ensemble, which consists in establishing the nature of the three dimensions and the boundaries of each component, leads *per se* to a reliable output, in the sense of either establishing that a molecule is therapeutically useful when it is, or that it is not therapeutically useful when it is not. The 'reliability-conferring' properties of a well-constructed intervention ensemble need to be specified on a case-by-case basis, but they certainly relate to managing uncertainties and providing clinical evidence. Therefore, a well-assembled intervention ensemble will give us reasons to believe that it is reliable, thereby showing that a molecule is, indeed, therapeutically useful (or that it is not). By strengthening the associations between an intervention and a clinical end-point (e.g., by finding the right co-interventions and optimal values, and addressing issues of confounding factors), an intervention ensemble will manage the uncertainties introduced by associationism (p1). Understanding the specific clinical context and finding optimal values/boundaries within which the molecule will be effective will sidestep the need for any information about the mechanism of action to warrant the use of a given drug (p3), and will avoid the pitfalls of the incompleteness and uncertainties of speculations about mechanisms of action[9]

---

[9] In our understanding, this view is typical of movements like Evidence-Based Medicine (EBM), which downplays the importance of mechanistic evidence. While EBM is certainly a major player in shaping the epistemology of

(Howick 2011). Finally, while pharmaceutical theory might be useful at various junctures of clinical translation (Aronson et al 2018; Kimmelman and London 2015), the justification of the results of a RCT are not dependent on theory or domain knowledge, i.e., reliability can be established despite atheoreticity (p2). All in all, building an intervention ensemble by following the proper vetted standards will provide good reasons to rely on the outputs of the process of clinical translation.

We call the second sense of reliability conferred by intervention ensembles *use-reliability*. Having built a robust intervention ensemble is not enough to guarantee that the molecule will be therapeutically useful in a new context. For instance, just knowing that a drug x is efficient in treating the condition y is not enough to safely prescribe x to treat all instances of y. However, the information gathered by constructing x's intervention ensemble can be used to assess whether the conditions under which x is shown to treat y are similar to the conditions of the context in which one wants to prescribe the drug. Recent calls to report protocols of trials such as SPIRIT or CONSORT (Chan et al 2013; Schulz et al 2010), which are typically taken to put one in a position of evaluating ensemble-reliability, are meant to also improve 'use-reliability'. In other words, information contained in a well-constructed intervention ensemble will give us reasons to believe that the pharmaceutical will work properly in a context that is similar to the original context of the intervention ensemble.

Our claim is that what is 'generative' is exactly the process of constructing reliable intervention ensembles. It thereby generates a research strategy that seeks to manage the challenges p1-3 through this process. Returning to the example in section 2.2, this is a similar transferal of an approach (here assembling an ensemble) to the transferal of the coding strategy to DNA. However, given the differences between the medical and the ML contexts, we cannot use the same exact procedures, nor even the idea of intervention ensemble as-is. In the next section, we will show how this work of adaptation can be carried out.

---

medicine, there are also other scholars who stress the importance of mechanisms, thereby arguing in favor of a more inclusive evidential pluralism (see for instance, Parkinnen et al 2018).

## 4 USING THE GENERATIVE ANALOGY BETWEEN CLINICAL TRANSLATION AND MACHINE LEARNING

Having introduced intervention ensembles (Section 3), we can use the analogy between ML and clinical translation to generate similar, but not identical, epistemic and methodological warrants (understood broadly) to mitigate risks associated with p1-p3 for ML and increase its reliability. Our proposal (and, we claim, it is implicitly London's proposal) is that we should take inspiration from intervention ensembles to generate an equivalent ensemble-template for the ML context. We call the ML equivalent *learning ensemble* (LE, see Figure 1). Ideally, a LE will confer both ensemble-reliability as well as use-reliability (section 3.3).

We generally define a learning ensemble LE as a set of components c1, c2, …, cn that (within the boundaries b of specific conditions sc1, sc2,...,scn) will provide reasons to believe that a ML system S is *ensemble-reliable*. If those conditions apply in a new context, and they are within the defined boundaries, then we have reasons to believe that S will be reliable in the new context as well (*use-reliability*). By increasing the disposition for ensemble-reliability and use-reliability, LE will also manage the risks associated to p1-3 (though without eliminating p1-3).

### 4.1. Components, dimensions, and reliability of learning ensembles

Using the generative analogy (section 2) leads us to believe that there may be similar dimensions to LEs as there are to ensembles in the clinical context. Furthermore, as in the case of intervention ensembles, one can distinguish between components within the dimensions. These components and dimensions constitute the information to meaningfully evaluate the system's reliability and to determine whether a certain ML system will reliably work in a new context.

In the clinical context, the dimensions of ensemble comprise information on the treatment, the population, and the outcomes. In order to identify the dimensions of LEs (and their components), we take inspiration from recent initiatives in medical AI that are aimed at improving how practitioners report the way ML systems in medicine are constructed: SPIRIT-AI (Cruz-Rivera et al 2020), CONSORT-AI (Liu et al 2020), and TRIPOD+AI (Collins et al 2024). These reporting initiatives are 'AI extensions' (as the titles of the SPIRIT-AI and CONSORT-AI articles suggest) of reporting guidelines for randomized controlled trials and computational models in medicine. These reporting guidelines provide the necessary information (i) to evaluate whether ML systems have been designed according to vetted standards, which are viewed as conducive to a reliable performance, and (ii) to implement ML systems in new contexts in such a way that they will perform reliably. Therefore, we can ascribe to these initiatives aims of improving both ensemble- and use-reliability. Drawing on these initiatives, we identify three dimensions of LEs. These are:

1. the '*boundaries of reliability*' *dimension*, i.e., the specific circumstances in which a ML system has been built, the methodological steps followed, and the justification for these steps
2. the *performance dimension*, i.e., the various metrics used to evaluate the performance of the ML system, and the justification for using such metrics
3. the *functional dimension*, i.e. the intended uses of the ML system, the evidence that the intended uses are achievable, and the relation of these uses to domain knowledge and domain norms

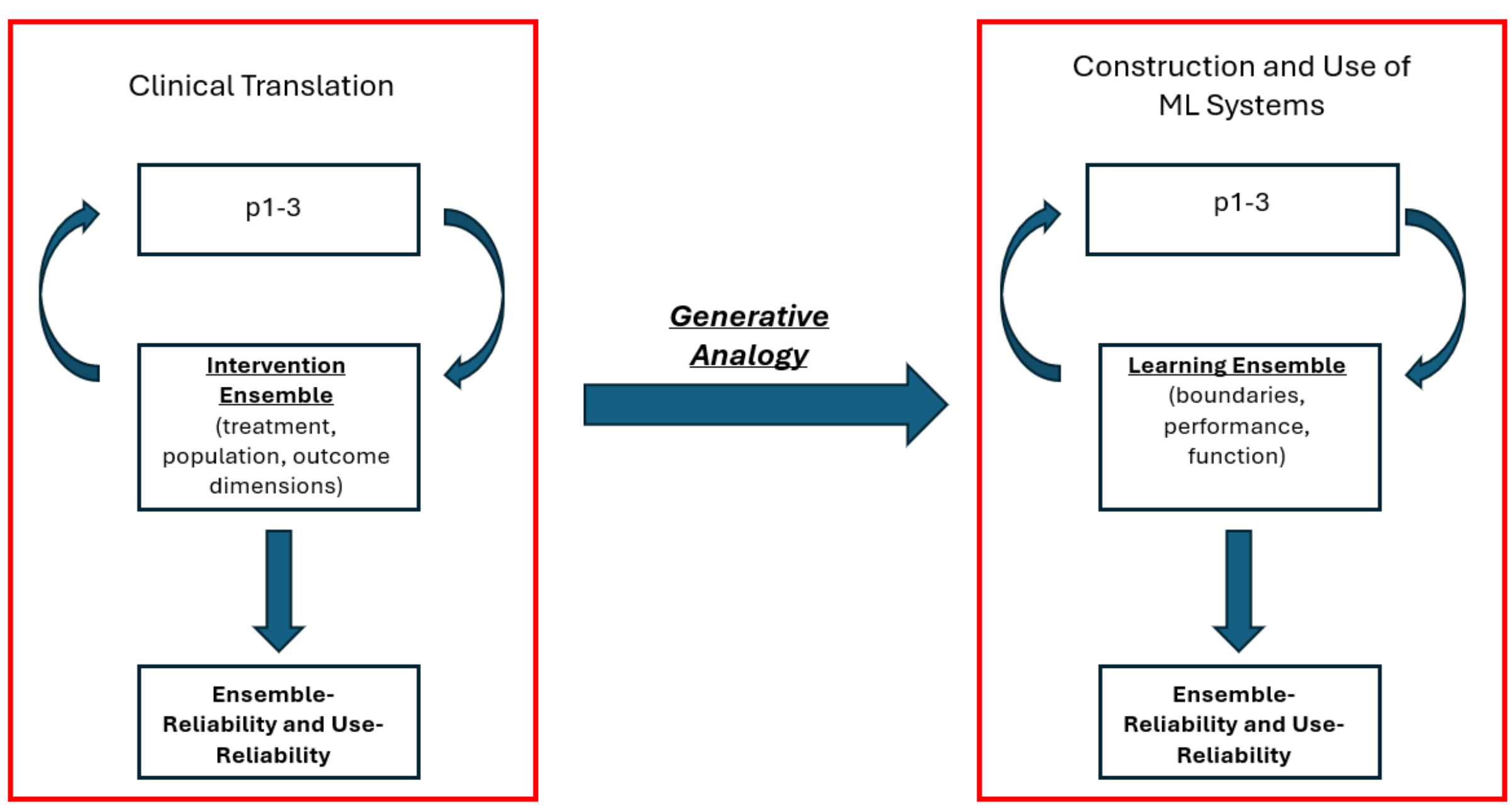

Figure 1. Generative analogy between clinical translation and machine learning

The components within each dimension will increase either ensemble-reliability or use-reliability of ML systems (or both). As a reminder, we intend 'ensemble-reliability' as the system's disposition to generate adequate outputs consistently in its original context, and use-reliability as the system's disposition to generate adequate outputs consistently in a new context. But what are the mechanisms through which each component within each dimension gives us reasons to believe that the LE is both ensemble-reliability and use-reliability? There are two main mechanisms[10]:

- positive mechanism: a given component will provide practitioners reasons to believe that the ML system has a disposition to produce adequate outputs, or that the ML system will deliver adequate outputs in a new (similar) context
- negative mechanism: a given component will provide reasons to believe that known failures that are known to have impacted both ensemble-reliability and use-reliability have been avoided. It is important to point out that negative mechanisms provide grounds to believe that a given ML system is reliable not *per se*, but rather that it is more reliable than it would be without that component avoiding known failures.

To sum up, the idea is that the components of different dimensions give us reasons to believe that the LE is ensemble-reliable and use-reliable by means of positive and negative mechanisms that, together, will increase practitioners' confidence to believe that outputs of ML systems are adequate (in the sense specified above), and known failures have been avoided. In this way, risks raised by p1-3 are mitigated, but without in fact eliminating p1-3 (because p1-3, as we specified above, characterize the way ML systems work). What our analysis will show is that establishing that a given ML system generates adequate outputs is a much more holistic (understanding this term 'informally') endeavor than just looking at brute quantitative performance metrics.

In the next sections, we will outline how different components of the three dimensions can potentially give practitioners reasons to believe that a given LE is ensemble-reliable and/or use-reliable (summarized in Table 1 at the end of Section 4). We are not providing a comprehensive and exhausting list of components and their reliability-conferring properties. Rather, we will exemplify some properties of LEs for MLs. In particular, for each component we will specify three things. First, we provide a description of the component. Second, we describe how the component is conducive of either ensemble-reliability or use-reliability (or both), and through which mechanism (either positive or negative, or both). Finally, where relevant we show the advantages

[10] We call these 'mechanisms', where this word should be intended in a broad sense, ranging from simple indicators to fully-fledged decision-making procedures.

of each component (or the disadvantages of not having the component) through well-known examples.

### *4.1.1 The Boundaries of Reliability*

The first dimension of LEs contains information on what we call 'the boundaries of reliability' (Table 1). A ML system is typically built to perform well in specific circumstances. We understand 'circumstances' (as it will be apparent below) in a broad sense. To make the case for ensemble-reliability, circumstances that allow an optimal performance need to be established, made explicit, and justified. To make the case for use-reliability, it has to be possible to compare the circumstances of implementation to the original circumstances in which the ML tool has been built. In the following, we discuss some examples of components within this dimension.

First, a given ML system will work reliably if used by a specific user group. This may be for different reasons e.g., the ML system is designed in such a way that its output or its functioning can be properly understood only by individuals with certain competencies. This is especially relevant for use-reliability. For instance, SPIRIT-AI (extension 11a iv, but see also TRIPOD-AI item 27b) stresses that poor clarity on how to use the human-AI interface may lead to confusion about "whether an error occurred due to a human deviation from the instructed procedure, or if it was an error made by the AI system" (p. 1359). In both cases, use-reliability is conferred via a negative mechanism: knowing that a given ML system can be handled properly only by individuals with a certain expertise, will narrow down the number of contexts in which the ML system's outputs will be leveraged competently, and will provide information as to which contexts to avoid.

A second class of components should specify the infrastructure supporting the ML system. SPIRIT-AI and CONSORT-AI (briefly) mention aspects concerning the specifications of suitable hardware or software. These aspects are essential to the ensemble-reliability as it needs to be ensured that the ML system has been developed and tested on well-functioning computational infrastructures, platforms, and reputable software. If a ML system has been constructed on the basis of dubious computational infrastructures, then this may raise concerns about its reliability. This is a mechanism with both positive and negative aspects. Reputable and well-functioning infrastructure will give us reasons to believe that the ML system is properly executing its computational and coding processes, while using dubious software or hardware will give us reasons to believe that errors might happen: it is well-known that silent degradation of hardware and software, while being difficult to be detected, might lead to a number of performance 'failures', including so-called, 'fail-partial', 'fail-transient', and 'fail-slow' behaviors (Gunawi et al 2018). A ML system that has been built by following vetted standards addressing these issues is more reliable than one that it has not. Similarly, this component is important to ensure use-reliability: if a ML system has been designed to perform well on specific hardware or software, then in a novel context of implementation comparable hardware and compatible software must be used.

One of the most crucial components of this dimension of LEs is input data. To evaluate ensemble-reliability and use-reliability, there are both positive and negative mechanisms related to the components specifying the characteristics of input data.

Let us start with the positive ones. Take data specification. Both TRIPOD+AI (Collins et al 2024) as well as SPIRIT-AI (Cruz Rivera et al 2020) emphasise the importance of reporting crucial information about the data used to train the ML system, e.g., sources of such data, rationale for using this data, and information on its representativeness. Ensemble-reliability is increased by, e.g., diversifying data provenance (e.g., multiple hospitals, multiple equipment types, etc). This provides more reasons for practitioners to believe that the ML system has learned stable patterns during training and validation, rather than idiosyncrasies of a single source. In terms of use-reliability, the representativeness of data is central, as described in TRIPOD+AI (item 5). This kind of documentation will provide practitioners reasons to believe that the ML system possesses a disposition to provide correct outputs given a certain task, in the sense that it has learned associations that generalize over a range of contexts including the one where the ML system is being deployed after training. It is important to point out that in both cases p1 and p2 are not eliminated. But while the ML systems still provide associations that remain theoretically unjustified (in the sense that the ML system does not provide any justification), our confidence that outputs are correct has increased.

Negative mechanisms also play a central role, both for ensemble-reliability and use-reliability. For instance, for ensemble-reliability it is important to specify protocols for the minimum quality-requirements for the input data, e.g., as stressed by SPIRIT-AI (extension 10 ii). Given this information, it is possible to evaluate whether the input data is of sufficient quality to avoid the well-known problem of 'garbage-in/garbage-out'. High-quality data can diminish issues of shortcuts, because such data can be curated to eliminate artifacts which might lead to shortcuts in the first place. High-quality data can also be used to diminish the nefarious consequences of adversarial attacks during adversarial training (Dong et al 2020). All this information gives practitioners reasons to believe that known failures have been properly addressed, and that the LE will be more 'ensemble-reliable' than a LE without those components. In order to illustrate what can go wrong when these components are not part of a LE, consider the case of AI used for detecting COVID-19 in chest radiographs (DeGrave et al 2021). In this study, it has been shown that existing ML systems relied on confounds which do not correlate with COVID-19, and that "probably reflect dataset-level differences in patient positioning and radiographic projection" (p. 614). Diverse data sources and more clarity on data specification could have decreased the chances of such shortcuts. But please note that these 'guardrails' do not eliminate p1-3; rather, they provide grounds for confidence despite them.

Information about input data is also crucial as a negative mechanism for evaluating use-reliability. At a basic level, information about data will suggest issues of generalization via possible mismatches between the training data distribution and the data distribution of the context of implementation (Freiesleben and Grote 2023), which can result in unreliable performance of the ML system in the novel context. In other words, knowing whether there is a distribution mismatch will provide reasons to believe that known failures of distribution shifts are properly avoided. There are also other aspects of distribution mismatches that can be identified by looking at data. For instance, one can identify the extent to which given input data is a proxy-measure or the provenance of input data, such as it being unprocessed or vendor-specific post-processing data (SPIRIT-AI, extension 10 ii). This type of information is useful to establish whether proxies are acceptable to the new context of implementation, or whether the processing procedures of a specific vendor are not compatible with the data used in the deployment context. Furthermore, knowing that a certain ML system is continuously retrained with data of a specific provenance can be important to anticipate possible natural distribution shifts or performativity in the novel context of implementation (FDA 2019; TRIPOD+AI item 12f; Freiesleben and Grote 2023; SPIRIT-AI, extension 11a iii). These are all negative mechanisms: with this information in hand, practitioners can avoid known failures. These components are particularly relevant to manage p1 and p3: natural distribution shifts are, indeed, risks related to p1-p3. And while these negative mechanisms do not eliminate p1 and p3, they provide a way *to manage* those risks.

One case of monumental failure of an unreliable LE (especially in terms of ensemble-reliability) caused by poor attention to the data-related component of the dimension 'boundaries of reliability' is the study by Poore et al (2020). Through a large-scale analysis of *The Cancer Genome Atlas* data, Poore et al's ML system detected distinctive microbial signatures in 32 types of cancer, which could be used to distinguish between tumor and normal tissues. In other words, they claimed that one could detect cancer (with an accuracy ranging from 95% to 100%) just by looking at the microbiome. Poore et al classified 7.2% of raw reads from human cancer genomes as non-humans on the basis of problematic alignments to the *Reference Human Genome*, and then found that these sequences matched sequences present in a genomic database of microorganisms. However, two important pieces of information about data input could have raised questions concerning ensemble-reliability, and in fact led to another study (Gihawi et al 2023) which invalidated the results. These 'clues' can be seen as instances of negative mechanisms. First, the fact that the alignment of the 7.2% of reads was difficult does not necessarily mean that reads are non-humans: cancer genomes are highly-mutated and they differ substantially from the *Reference Human Genome* which, it is important to remind, is an idealized consensus genome that, technically speaking, is the genome of no one. This means that it is not unreasonable to believe that those reads could be, in fact, human reads. Therefore, having information about what the data are about might lead one to think that the 'misalignment' to the *Reference Human Genome* could simply be a mistake. Second, as reported by Gihawi et al (2023), it is well known that genomic databases (including databases of microorganisms) are contaminated with large numbers of mislabeled sequences which are, in fact, sequences belonging to human beings. This problem of cross-species contamination is large-scale but, again, well-known (and that is why algorithms for alignments are conservative). This means that those reads that were classified as belonging to given microbial species could be, in fact, simply sequences that contaminated the genomic data of microorganisms, and assembled erroneously into one single consensus sequence. With this information about the nature of data input, another negative mechanism emerges: there are reasons to believe that the signatures detected by the ML system are *not signatures at all*. In fact, Gihawi et al shows that this was the case (together with other things to be discussed below).

*4.1.2 Performance Criteria*

The first dimension provide information on the conditions and the boundaries within which a ML system is thought to have a disposition to provide adequate outputs or avoid known failures. However, it does not contain information about actual measures of the performance of the ML system, i.e. a measure of the system's record of providing adequate output, as defined above. Quantitative metrics used to measure the performance of a given ML system are specified by what we call the *performance* dimension. These measures of performance need to be contextualized within the first dimension and, as we will see, within the third one as well.

There exists a large body of literature on measuring performances of ML systems. The components of this dimension overlap significantly with the reliability indicators of technical performance of algorithms in computational reliabilism (Duran 2026). Duran explicitly mentions validation methods, showing how a judicious choice of specific validation methods and their results provide "a good indication of the algorithm's accuracy and margin of error" (p. 66). Quantitative measures of performance are important to evaluate ensemble-reliability, and they leverage both negative and positive mechanisms. This is particularly apparent in how the performance of, say, ML supervised systems is measured, which is typically through a loss function, namely the distance between the correct prediction and the prediction generated by the ML system (Termine et al 2026). This is technically a negative mechanism (i.e. it measures the error), but at the same time it is perceived as giving a measure of how well the system performs. Assuming that the original context of the ML system and the new context in which the ML system is intended to be used are comparable, technical performance should also provide to practitioners reasons to believe that the ML system will perform adequately in the new context: it is relevant for use-reliability. Another important (though underplayed by the reporting initiatives considered) aspect is that ML systems should also report the level of uncertainty for a certain output (Grote 2021). Reporting uncertainties is a case of positive mechanism, for both ensemble-reliability and use-reliability: it suggests how confident is a ML system of the strength of the association between an input and an output. Additionally, integrating the output with the use of Explainable AI (XAI) tools can, in part, provide a negative mechanism to help with both ensemble-reliability and use-reliability (this is another item surprisingly missing from the reporting initiatives). For instance, XAI tools, while not providing positive evidence for something to work (Ratti 2022), can nonetheless flag cases of shortcuts (and hence mitigates risks associated to p1 and p3), by showing that a ML system has indeed leveraged, e.g., pixels of an image that are *unrelated* to the particular clinical condition.

An important component is the performance for key subgroups (TRIPOD-AI item 23e; CONSORT-AI item 18). What a good LE should provide are results of performance as stratified by (e.g. in medicine) demographics, disease subtypes, severity levels, etc. This leverages both positive and negative mechanism. Because 'subtypes' provides information for the scope of performance, then the ML system performance within each subtype provide a specific measure of ensemble-reliability for that subgroup, as well as providing confidence for how well the ML system (again, for that subgroup) will do for use-reliability in a new context. By giving estimates for each subgroup, it will also rule out poor-performance for specific subgroups, thereby impacting grounds for both ensemble-reliability, and use-reliability. In particular, the error that is mitigated with subgroup performance analysis is, notoriously, *algorithmic bias*. Problematic cases are easy to find. For instance, Seyyed-Kalantari et al (2021) found that models based on three available radiology datasets performed poorly on members of underserved populations. Providing a subgroup performance analysis can help to mitigate risks associated with algorithmic biases. While subgroup documentation does not solve issues of p1 or p3, it still provides grounds for reliability claims in the way indicated.

Further information relevant to determine ensemble-reliability are justifications of how and why a given performance metric has been chosen. For instance, any metric is subjected to various tradeoffs between false-positives and false-negatives, in particular the precision-recall tradeoff and the sensitivity-specificity tradeoffs. Precision is intended to reduce false positives, while recall aims to reduce false negatives. Similarly, specificity is intended to reduce false positives, while sensitivity aims to reduce false negatives. The difference is that the pair sensitivity-specificity is focused on both positive and negative classes, while precision-recall ignores true negatives altogether. Ratti and Graves (2022) notice that, depending on the goals of the ML system, using one pair rather than another might make a difference as to how reliable the ML system will be. For instance, if the purpose of the system is diagnostic, then sensitivity-specificity may be a better tradeoff to engage with. This is because both positive and negative classes (disease vs not-disease) are a concern of practitioners.  But if the system is retrieving disease information from, say, patient records, then precision-recall might be a better choice, because retrieving information from an electronic health record faces extreme imbalance (i.e., most entries in the records will lack mentions of the information required), which can be avoided by using precision-recall. Documenting these aspects can help with ensemble-reliability, as it can point to wrongly-used metrics (i.e., a negative mechanism), thereby making practitioners question performance.

Information on how trade-offs have been addressed is important for use-reliability. This is for a particular negative mechanism: if a user thinks that in the new context of implementation false positives should be avoided more than false negatives, then knowing how the tradeoff between false positive and false negative has been treated in the ML will provide crucial information to decide whether to use the ML system in the novel context. In other words, information on how tradeoffs of performance have been set, prevents practitioners from using a ML system that will fail to achieve a given goal in the new context of implementation.

### *4.1.3 The Functional Dimension*

Philosophers of technology have drawn a distinction between *effect function* and *purpose function* of technical artifacts (Vermaas 2009; van Eck 2015). On the one hand, effect functions designate the desired effect of the behaviour of a technical artifact. In the case of an electric screwdriver, the effect function will be to tighten or loose screws. On the other hand, purpose functions designate those state of affairs in the real-world that effect functions are taken to contribute to. In the case of the electric screwdriver, the purpose function might be to hang a painting. There can be misalignments between effect functions and intended purpose functions, in the sense that effect functions are not conducive of purpose functions. Drawing on this distinction, we think that additional information has to be provided to make sure that the ML system not only performs well (i.e., the second dimension of LEs) within a specific set of conditions (i.e., the first dimension of LEs), but also to make sure that its purpose function is well-specified, that it is aligned with the context of use, and that there is evidence that the purpose function is indeed facilitated by the system's effect function. We call this third dimension the *functional* dimension (see Table 1).

In general, the functional dimension increases the reliability of the ML system by providing reasons to believe that the ML system's function is appropriate for the intended use and the context of use - in other words, that a given ML system is the right tool for the task at hand. The main components are the relevant effect and purpose functions. ML systems' effect functions are typically predictions or classifications. In contrast, the purpose functions to which ML systems should contribute may vary. In the case of medical AI, the FDA has recently (2019) stressed the importance of specifying how the ML system is intended to contribute to healthcare practices, e.g. whether it should be used to diagnose or to drive (rather than to inform) clinical management. For instance, if the purpose function is to provide a diagnosis in such a way that the ML system trumps all further reasoning, then we might reasonably require performance metrics to be much more stringent, given the potential consequences. In other words, specifying the purpose function can provide further information to contextualise performance and, ultimately, to evaluate ensemble-reliability. Specifying the purpose function is important also for use-reliability. For example, if a ML system has been developed to inform clinical management then there is no reason to assume that it can be used reliably to drive clinical management. The importance of the functional dimension is underscored in the reporting initiatives mentioned above. TRIPOD+AI (items 8a and 8b) alludes to the importance of providing clear outcomes and instructions on how to interpret them. SPIRIT-AI (extension 11a) is even more explicit in requesting clarity on how the ML system will contribute to specific decision-making procedures or clinical practices, and on how to interpret its outputs. For instance, in the case of a skin cancer detection system producing a percentage likelihood as output, SPIRIT-AI guidelines point out that what can possibly acted upon the outcome must be clearly specified, such as the intended pathway and the threshold for the intended pathway (e.g., lesion excision if positive, with probability of, say, 80%).

In order to understand the importance of the functional dimension of LEs, let us be more explicit on what can go wrong when the components of this dimension are not scrutinized seriously. In particular, there might be three problems with this dimension:

1. the purpose function is misleading
2. there is a misalignment between the effect function and the context of the purpose function
3. The effect function is unable to contribute to the achievement of the purpose function

In the first case, if the purpose function of a ML system developed in, say, a scientific context is completely at odds with the norms and domain-knowledge of that context, then its reliability is jeopardised. This is because, while the ML system might perform well (i.e., the second dimension of LEs), the outputs themselves might be misleading or based on poorly supported assumptions. Therefore, a purpose function that is well aligned to a given scientific context, will give us reasons to believe that outputs will not be inconsistent with domain knowledge and norms. Duran (2026) calls this reliability indicator ‘knowledge-based integration’, and he shows the perils of misleading purpose functions with an example taken from facial recognition for crime detection. In this case, the effect function is the classification of images, and the purpose function is detecting criminality via images of faces. An obvious negative mechanism emerges when considering the purpose function. In particular, the problem with the purpose function is that the categories this AI system “purports to use (...) are posited in isolation from established evidence, models of criminal psychology, social studies of crime, and the relevant theories of criminality” (p. 71). In other words, the purpose function posited by practitioners should be rejected on the basis of theories of crime. By identifying the extent to which *the purpose function* of the ML system is aligned with domain-knowledge and domain norms, one can evaluate both ensemble-reliability and use-reliability, at least via this negative mechanism.

The second case is more subtle but potentially no less catastrophic. For instance, Mussgnug (2022) notices that in AI systems used to investigate poverty, framing the measurement of poverty in ‘predictive’ terms necessarily assumes, in the context of design, methodological assumptions about the validity and suitability of a given measurement, which are not aligned with considerations of validity and suitability of the context of implementation. This points to a tension between *effect function* (what the ML system does, its outputs) and the norms governing the context where the purpose function is envisioned to be realized: the tension will simply go unnoticed unless effect and purpose functions are explicitly documented. Another interesting failure of aligning effect-functions with norms governing the context of the purpose function is the work by Gihawi et al (2023) invalidating the study of microbiome signatures in cancer (Poore et al 2020), which we have discussed when describing the first dimension of LEs. In addition to the problems identified in Section 4.1.1, one significant problem was a misalignment between the effect functions and domain knowledge of the context of the purpose function. In this context, the effect function of ML systems are microbiome signatures in cancer (delivered via classifications), the purpose function is constructing a coherent picture of how cancer develops, and the context of the purpose function is cancer genomics and microbiome studies. The misalignment lies in the fact that many *genera* identified by the ML system had never been previously reported in the context of human disease, and some not even reported in humans (e.g., species associated with extreme, non-human, environments). These are all red flags, because the effect functions identify signatures that either have nothing to do with cancer, or nothing to do with human microbiome - i.e., the mismatch between outputs and domain-knowledge of the context of genomics give practitioners reasons to believe that the ML system is prone to false positives.

The third case is when the effect function cannot possibly contribute to the achievement of the purpose function. A classic example is the famous study by Caruana et al. (2016) on the ML system that ranked asthmatic patients as having a low probability of dying from pneumonia. In this case, effect functions are classifications of a patient as being high or low risk, the purpose function is triage, and the context of the purpose function is emergency medicine. The recommendation here should not be to opt for a more simple and explainable ML system (as Caruana et al. suggest); rather, it is to realize that the ML system, because it did not take into account the extensive treatments that asthmatic patients typically receive, could not reliably contribute to the purpose function (i.e., triage), which is typically "to optimize the allocation of medical resources against a baseline risk of death that is *independent* [emphasis added] of current medical practice" (London 2019, p. 19). Therefore, documenting that the data used *indeed* reflected probability of death *given* the current medical practice of aggressively treating asthmatic patients in ICU, could have shown that the classifications prompted by the ML system were unable to achieve the purpose function - documenting all these aspects is a classic case of negative mechanism increasing the correctness of a diagnosis of ensemble-reliability. Please note that all of this has nothing to do with the *atheoreticity* of the way the algorithm learns or the model (that is, p2); it is about the relations between the purpose of a ML system (which can and should be made entirely transparent) and the domain knowledge and domain norms of the context of implementation.

| DIMENSION | COMPONENT | MECHANISM (NM = negative; PM = positive) | EXAMPLE (if applicable) |
|---|---|---|---|
| Boundaries of Reliability | User-Group Specification | Use-Reliability: NM | Misuse of ML Systems |
| | Hardware and Software | Ensemble-Reliability: PM and NM. Use-Reliability: NM | n/a |
| | Data Provenance | Ensemble-Reliability: PM. Use-Reliability: PM, NM | Shortcuts in COVID-19 ML System; Microbiome Case |
| | Data Quality | Ensemble-Reliability: NM. Use-Reliability: NM | Shortcuts in COVID-19 ML System |
| | Data Specification | Ensemble-Reliability: NM | Microbiome Case |
| Performance | Performance Measures | Ensemble-Reliability: NM and PM | n/a |
| | Reporting Uncertainties | Ensemble-Reliability: PM; Use-Reliability: PM | n/a |
| | XAI Tools | Ensemble-Reliability: NM | Cases of Shortcuts |
| | Subgroups Performance | Ensemble-Reliability: PM and NM; Use-Reliability: PM and NM | ML System Based on Radiology Datasets Performed Poorly on Underserved Populations |
| | Justification of Metrics | Ensemble-Reliability: NM | Precision-recall vs Sensitivity-Specificity |
| | Justification of Tradeoffs | Use-Reliability: NM | n/a |
| Function | Justification of Purpose Function | Ensemble-Reliability: NM | Face-Recognition ML System to Predict Criminality |
| | Relation Between Effect Functions and the Context of the Purpose Function | Ensemble-Reliability: NM | ML in Poverty Studies; Microbiome Case |
| | Ability of Effect Function to Achieve Purpose Function | Ensemble-Reliability: NM | Health-risk Assessment ML System and Ashmatic Patients |

Table 1. Summary of Dimensions and Components of a LE

## 5 CONCLUSION

In this article, we have provided a philosophical interpretation of the parallels between AI and medicine, and have used these parallels for developing a reliabilist framework for ML. We have interpreted the parallels as analogies (section 2). In particular, we have shown that the process of clinical translation can be understood as a process of assembling intervention ensembles, for which ensemble- and use-reliability are assessed (section 3). Interpreting the analogy with AI as generative, we used it to suggest the construction of an equivalent ensemble (i.e., a LE) to assess ensemble- and use-reliability of ML systems (section 4). Thereby, we have distinguished three dimensions of LEs: boundaries of reliability; performance; and functionality.

Our account of ML reliability is compatible with other accounts present in the literature, in particular with computational reliabilism (Duran and Formanek 2018; Duran 2026). With computational reliabilism, we share the idea that an epistemology of transparency is unsuitable for evaluating ML systems. And even if it were suitable, our framework shows that ‘transparency’ is redundant. This is because an epistemology of transparency will claim to solve p1-3 in order to manage their risks, while our account (as well as computational reliabilism) does not need to solve p1-3 in order to manage their risks. But our account differs from computational reliabilism in important respects. First, computational reliabilism is an epistemology of algorithms and it casts a wide net encompassing computer simulations, machine learning, GOFAI, etc. Our account only addresses the construction of LEs: it is restricted to ML and, as a result, the structure of LEs really reflects ML specificities. As a consequence of the first aspect, the dimensions and components of LEs and reliability indicators as conceptualized by Duran overlap, but do not coincide. Second, our account puts significant emphasis on how reliability-conferring properties of LEs components indeed confer reliability, while computational reliabilism does less so for reliability indicators. Finally, our distinction between ensemble-reliability and use-reliability is not, to our knowledge, reflected in computational reliabilism.

Despite the wealth of details provided, only some examples of relevant components of LEs are discussed here, and much more needs to be said about how the construction and evaluation of LEs should look like. As Grote et al (2024) notice, building a robust and reliable ML system "typically involves a trial-and-error process, requiring a combination of domain knowledge, external evaluation with out-of-distribution data, data augmentation, explainable AI techniques, and continuous retraining" (p. 5). This article, though, paves the way to an epistemology of LEs, and provides a framework that can guide LEs construction and evaluation.

**Acknowledgements**. Thanks to Juan Duran and Richard Pettigrew for useful comments on an earlier version of this manuscript.

## REFERENCES

Alvarado, R., & Humphreys, P. (2017). Big data, thick mediation, and representational opacity. *New Literary History*, *48*(4), 729–749. https://doi.org/10.1353/nlh.2017.0037

Aronson, J. K., la Caze, A., Kelly, M. P., Parkkinen, V. P., & Williamson, J. (2018). The use of mechanistic evidence in drug approval. *Journal of Evaluation in Clinical Practice*, *24*(5), 1166–1176. https://doi.org/10.1111/jep.12960

Bartha, P. (2022). Analogy and analogical reasoning, *Stanford Encyclopedia of Philosophy*

Boge, F. J. (2022). Two Dimensions of Opacity and the Deep Learning Predicament. *Minds and Machines*, *32*(1), 43–75. https://doi.org/10.1007/s11023-021-09569-4

Chan, A.-W., Tetzlaff, J. M., Altman, D. G., Laupacis, A., Gøtzsche, P. C., Krleža-Jeric, K., Hró bjartsson, A., Mann, H., Dickersin, K., Berlin, J. A., Doré, C. J., Parulekar, W. R., Summerskill, W. S., Groves, T., Schulz, K. F., Sox, H. C., Rockhold, F. W., Rennie, D., & Moher, D. (2013).

SPIRIT 2013 Statement: Defining Standard Protocol Items for Clinical Trials. In *Ann Intern Med* (Vol. 158). www.annals.org

Collins, G. S., Moons, K. G. M., Dhiman, P., Riley, R. D., Beam, A. L., van Calster, B., Ghassemi, M., Liu, X., Reitsma, J. B., van Smeden, M., Boulesteix, A. L., Camaradou, J. C., Celi, L. A., Denaxas, S., Denniston, A. K., Glocker, B., Golub, R. M., Harvey, H., Heinze, G., … Logullo, P. (2024). TRIPOD+AI statement: Updated guidance for reporting clinical prediction models that use regression or machine learning methods. *BMJ*. https://doi.org/10.1136/bmj-2023-078378.

Cruz Rivera, S., Liu, X., Chan, A. W., Denniston, A. K., Calvert, M. J., Darzi, A., Holmes, C., Yau, C., Moher, D., Ashrafian, H., Deeks, J. J., Ferrante di Ruffano, L., Faes, L., Keane, P. A., Vollmer, S. J., Lee, A. Y., Jonas, A., Esteva, A., Beam, A. L., … Rowley, S. (2020). Guidelines for clinical trial protocols for interventions involving artificial intelligence: the SPIRIT-AI extension. *Nature Medicine*, *26*(9), 1351–1363. https://doi.org/10.1038/s41591-020-1037-7

DeGrave, A. J., Janizek, J. D., & Lee, S. I. (2021). AI for radiographic COVID-19 detection selects shortcuts over signal. *Nature Machine Intelligence*, *3*(7), 610–619. https://doi.org/10.1038/s42256-021-00338-7

Djulbegovic, B. (2007). Articulating and responding to uncertainties in clinical research. *Journal of Medicine and Philosophy*, *32*(2), 79–98. https://doi.org/10.1080/03605310701255719

Dong, Y., Deng, Z., Pang, T., Zhu, J., & Su, H. (2020). Adversarial Distributional Training for Robust Deep Learning. *NeurIPS 2020*.

Durán, J.M. (2026). Beyond Transparency: Computational Reliabilism as an Externalist Epistemology of Algorithms. In: Durán, J.M., Pozzi, G. (eds) Philosophy of Science for Machine Learning. Synthese Library, vol 527. Springer, Cham.

Durán, J. M., & Formanek, N. (2018). Grounds for Trust: Essential Epistemic Opacity and Computational Reliabilism. *Minds and Machines*, *28*(4), 645–666. https://doi.org/10.1007/s11023-018-9481-6

Falconer, I. (2019). Vortices and atoms in the Maxwellian era. Philosophical Transactions A, 377, 20180451.

FDA. (2019). Proposed Regulatory Framework for Modifications to Artificial Intelligence / Machine Learning ( AI / ML ) -Based Software as a Medical Device ( SaMD ) - Discussion Paper and Request for Feedback. *U.S Food & Drug Administration*, 1–20.

Freiesleben, T., & Grote, T. (2023). Beyond generalization: a theory of robustness in machine learning. *Synthese*, *202*(4). https://doi.org/10.1007/s11229-023-04334-9

Genin, K., & Grote, T. (2021). Randomized Controlled Trials in Medical AI. *Philosophy of Medicine*, *2*(1). https://doi.org/10.5195/pom.2021.27

Gihawi, A., Ge, Y., Lu, J., Puiu, D., Xu, A., Cooper, C. S., Brewer, D. S., Pertea, M., & Salzberg, S. L. (2023). Major data analysis errors invalidate cancer microbiome findings. *mBio*, *14*(5). https://doi.org/10.1128/mbio.01607-23

Goldman, A.I. (1979). What is Justified Belief?. In: Pappas, G.S. (eds) Justification and Knowledge. Philosophical Studies Series in Philosophy, vol 17. Springer, Dordrecht. https://doi.org/10.1007/978-94-009-9493-5_1

Grote, T., Genin, K., & Sullivan, E. (2024). Reliability in Machine Learning. *Philosophy Compass*, *19*(5). https://doi.org/10.1111/phc3.12974

Grote, T. (2021). Trustworthy medical AI systems need to know when they don't know. In *Journal of Medical Ethics* (Vol. 47, Issue 5, pp. 337–338). BMJ Publishing Group. https://doi.org/10.1136/medethics-2021-107463

Gunawi, H. S., Suminto, R. O., Sears, R., Golliher, C., Sundararaman, S., Lin, X., Emami, T., Sheng, W., Bidokhti, N., Mccaffrey, C., Grider, G., Fields, P. M., Harms, K., Ross, R. B., Jacobson, A., Ricci, R., Webb, K., Alvaro, P., Runesha, H. B., … Li, H. (2018). Fail-Slow at Scale: Evidence of Hardware Performance Faults in Large Production Systems. *Proceedings of the 16th USENIXConference on File and Storage Technologies (FAST '18)*.

Hesse, M. (1966). *Models and Analogies in Science*. Notre Dame University Press.

Howick, J. (2011). *The Philosophy of Evidence-based Medicine*. John Wiley & Sons.

Humphreys, P. (2011). Computational science and its effects. In M. Carrier & A. Nordmann (Eds.), *Science in the Context of Application* (Boston Stu). Springer.

Kay, L. (2000). *Who wrote the book of life? A History of the Genetic Code*. Stanford University Press.

Kimmelman, J. (2012). A theoretical framework for early human studies: Uncertainty, intervention ensembles, and boundaries. In *Trials* (Vol. 13). https://doi.org/10.1186/1745-6215-13-173

Kimmelman, J., & London, A. J. (2015). The Structure of Clinical Translation: Efficiency,

Liu, X., Cruz Rivera, S., Moher, D., Calvert, M. J., Denniston, A. K., Chan, A. W., Darzi, A., Holmes, C., Yau, C., Ashrafian, H., Deeks, J. J., Ferrante di Ruffano, L., Faes, L., Keane, P. A., Vollmer, S. J., Lee, A. Y., Jonas, A., Esteva, A., Beam, A. L., … Rowley, S. (2020). Reporting guidelines for clinical trial reports for interventions involving artificial intelligence: the

CONSORT-AI extension. *Nature Medicine*, *26*(9), 1364–1374. https://doi.org/10.1038/s41591-020-1034-x

London, A. J. (2019). Artificial Intelligence and Black-Box Medical Decisions: Accuracy versus Explainability. *Hastings Center Report*, *49*(1), 15–21. https://doi.org/10.1002/hast.973

Mussgnug, A. M. (2022). The predictive reframing of machine learning applications: good predictions and bad measurements. *European Journal for Philosophy of Science*, *12*(3). https://doi.org/10.1007/s13194-022-00484-8

Parkkinen, V.-P., Wallmann, C., Wilde, M., Clarke, B., Illari, P., Kelly, M. P., Norell, C., Russo, F., Shaw, B., & Williamson, J. (2018). *Evaluating Evidence of Mechanisms in Medicine Principles and Procedures*, Springer

Ratti, E. (2022). Integrating Artificial Intelligence in Scientific Practice: Explicable AI as an Interface. *Philosophy and Technology* (Vol. 35, Issue 3). Springer Science and Business Media B.V. https://doi.org/10.1007/s13347-022-00558-8

Ratti, E., & Graves, M. (2022). Explainable machine learning practices: opening another black box for reliable medical AI. *AI and Ethics*. https://doi.org/10.1007/s43681-022-00141-z

Russo, F. (2023). What Can AI Learn from Medicine? *Digital Society*, *2*(2). https://doi.org/10.1007/s44206-023-00061-3

Schulz K F, Altman D G, Moher D. CONSORT 2010 Statement: updated guidelines for reporting parallel group randomised trials, *BMJ* 2010; 340 :c332 doi:10.1136/bmj.c332

Seyyed-Kalantari, L., Zhang, H., McDermott, M. B. A., Chen, I. Y., & Ghassemi, M. (2021). Underdiagnosis bias of artificial intelligence algorithms applied to chest radiographs in under-served patient populations. *Nature Medicine*, *27*(12), 2176–2182. https://doi.org/10.1038/s41591-021-01595-0

Stegmann, U. E. (2016). “Genetic Coding” Reconsidered: An Analysis of Actual Usage. In *Source: The British Journal for the Philosophy of Science* (Vol. 67, Issue 3).

Termine, A., Ratti, E., & Facchini, A. (2026). Machine learning and theory-ladenness: a phenomenological account. *Synthese*, *207*(3), 94. https://doi.org/10.1007/s11229-026-05454-8